\documentclass[runningheads]{llncs}

\usepackage{eccv}

\usepackage{eccvabbrv}
\usepackage{wrapfig}
\usepackage{pifont} 
\usepackage[table]{xcolor}
\usepackage{multirow}
\definecolor{cvprblue}{rgb}{0.21,0.49,0.74}
\usepackage{graphicx}
\usepackage{booktabs}

\usepackage[accsupp]{axessibility}  

\usepackage{hyperref}

\usepackage{orcidlink}

\begin{document}

\title{EventVGGT: Exploring Cross-Modal Distillation for Consistent Event-based Depth Estimation} 

\titlerunning{EventVGGT}




\author{Yinrui Ren\inst{1, 2}* \and
Jinjing Zhu\inst{1, 3}* $^\dagger$ \and 
Kanghao Chen\inst{1}*\and
Zhuoxiao Li\inst{1}* \and
Jing OU\inst{1} \and
Zidong Cao\inst{1} \and
Tongyan Hua\inst{1} \and
Peilun Shi\inst{3} \and
Yingchun Fu\inst{2} \and
Wufan Zhao\inst{1}$\nmid $ \and
Hui Xiong\inst{1}$\nmid $ \\ * Equal Contribution, $^\dagger$ Project Lead, $\nmid $ Corresponding author
}
\institute{Hong Kong University of Science and Technology (Guangzhou) \and 
South China Normal University \and
Chinese University of Hong Kong\\
}

\authorrunning{Yinrui Ren, Jinjing Zhu et al.}

\maketitle

\begin{figure}
    \centering
    \includegraphics[width=0.99\textwidth]{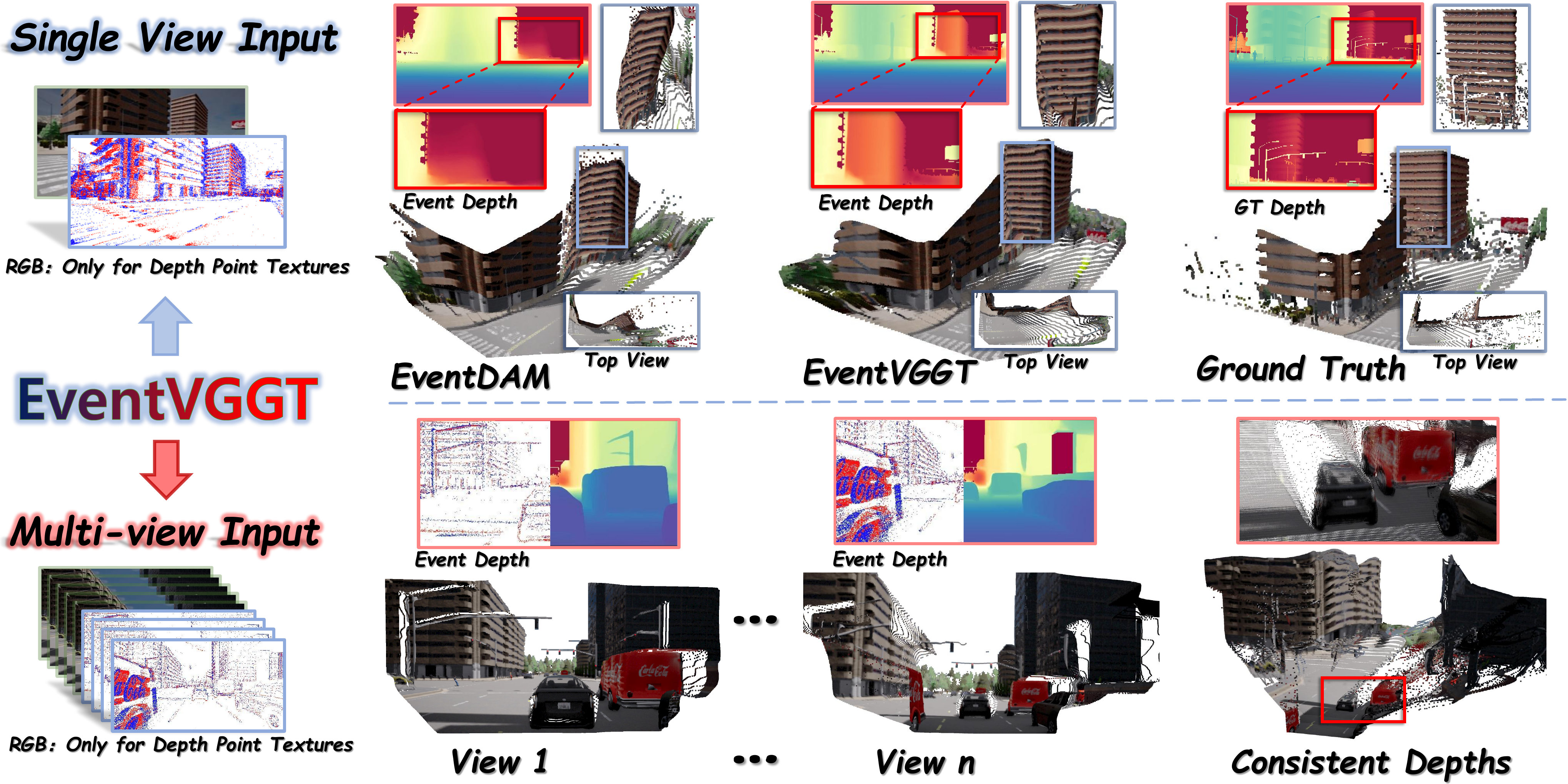}
        \captionof{figure}{We introduce \textbf{EventVGGT}, a novel framework that explicitly models asynchronous event streams as coherent video sequences. By distilling robust multi-view geometric priors from the Visual Geometry Grounded Transformer (VGGT), our approach achieves highly accurate and temporally consistent event-based depth estimation. \textit{Note}: Depth inference relies exclusively on event data; RGB images are used solely for colorizing the reconstructed 3D point clouds.}
        \label{fig:teaser}
\end{figure}

\begin{abstract}

Event cameras offer superior sensitivity to high-speed motion and extreme lighting, making event-based monocular depth estimation a promising approach for robust 3D perception in challenging conditions. However, progress is severely hindered by the scarcity of dense depth annotations. 
While recent annotation-free approaches mitigate this by distilling knowledge from Vision Foundation Models (VFMs), a critical limitation persists: they process event streams as independent frames. 
By neglecting the inherent temporal continuity of event data, these methods fail to leverage the rich temporal priors encoded in VFMs, ultimately yielding temporally inconsistent and less accurate depth predictions. 
To address this, we introduce \textbf{EventVGGT}, a novel framework that explicitly models the event stream as a coherent video sequence. To the best of our knowledge, we are the first to distill spatio-temporal and multi-view geometric priors from the Visual Geometry Grounded Transformer (VGGT) into the event domain.
We achieve this via a comprehensive tri-level distillation strategy: (i) \textbf{Cross-Modal Feature Mixture (CMFM)} bridges the modality gap at the output level by fusing RGB and event features to generate auxiliary depth predictions; (ii) \textbf{Spatio-Temporal Feature Distillation (STFD)} distills VGGT's powerful spatio-temporal representations at the feature level; and (iii) \textbf{Temporal Consistency Distillation (TCD)} enforces cross-frame coherence at the temporal level by aligning inter-frame depth changes. 
Extensive experiments demonstrate that EventVGGT consistently outperforms existing methods—reducing the absolute mean depth error at 30m by over 53\% on EventScape (from 2.30 to 1.06)—while exhibiting robust zero-shot generalization on the unseen DENSE and MVSEC datasets. The code is available at~\url{https://github.com/yinruiRen/EventVGGT}.

\end{abstract}

\section{Introduction}
\label{sec:intro}

Event cameras~\cite{gallego2020event, zheng2023deep} are bio-inspired sensors that asynchronously encode logarithmic intensity changes into sparse event streams. Unlike conventional RGB cameras, they offer exceptional temporal resolution and dynamic range, enabling robust perception under high-speed motion and challenging illumination conditions~\cite{liang2024towards, tulyakov2022time}. Consequently, they have driven significant advancements across various vision tasks, including object detection~\cite{li2022asynchronous, peng2023better}, tracking~\cite{chen2024emoe, gehrig2018asynchronous, zhang2021object}, and segmentation~\cite{jia2023event, jing2024hpl, chen2024segment, kong2024openess}. Among these, event-based monocular depth estimation~\cite{HidalgoCarrio2020LearningMD, devulapally2024multi, pan2024srfnet} has emerged as a critical capability for autonomous driving and robotic navigation~\cite{tulyakov2019learning, ranccon2022stereospike}. However, despite recent efforts to fuse the complementary strengths of event and RGB modalities~\cite{pan2024srfnet, hamaguchi2023hierarchical}, progress in event-based depth estimation remains bottlenecked by the scarcity of large-scale datasets with dense depth annotations.

To address this scarcity, recent methods~\cite{zhu2025depth, bartolomei2025depth} leverage Vision Foundation Models (VFMs) to generate high-quality pseudo-depth labels from paired RGB-event data, bypassing the need for ground-truth annotations. For instance, EventDAM~\cite{zhu2025depth} employs dense-to-sparse distillation on isolated event frames, while DepthAnyEvent~\cite{bartolomei2025depth} integrates temporal modules before transferring knowledge from a single-image teacher (DAM V2~\cite{yang2024depth}). Although these VFM-based approaches alleviate the annotation bottleneck, a critical limitation persists: \textit{by treating event streams as independent frames, they fail to leverage the inherent temporal continuity of the data and the temporal priors of VFMs, ultimately yielding temporally inconsistent and suboptimal depth predictions}.

Among recent VFMs, models equipped with multi-view geometric reasoning are ideal teachers, as their spatio-temporal 3D priors naturally align with the continuous dynamics of event streams. A prime example is the Visual Geometry Grounded Transformer (VGGT)~\cite{wang2025vggt}, which jointly infers depth, camera poses, and point clouds from multiple views. Unlike single-image baselines (\eg, DAM V2~\cite{yang2024depth}) that lack cross-frame awareness, VGGT implicitly enforces geometric consistency across views via alternating frame-wise and global attention. Inspired by this, we propose to adapt VGGT's powerful multi-view reasoning for consistent event-based depth estimation. By treating the asynchronous event stream as a dense sequence of views, we explicitly exploit VGGT to model the temporal constraints and geometric coherence inherent in event data.

To this end, we introduce \textbf{EventVGGT}, an annotation-free framework that treats the event stream as a continuous video sequence and distills rich spatio-temporal priors from a VGGT teacher to achieve consistent event-based depth estimation (see Fig.~\ref{fig:teaser}). To effectively transfer this knowledge, we propose a tri-level distillation strategy: (i) At the \textit{output level}, we introduce the \textbf{Cross-Modal Feature Mixture (CMFM)} module, which bridges the substantial modality gap by fusing RGB and event features to generate an auxiliary depth prediction. By explicitly supervising this auxiliary output with the teacher’s high-fidelity RGB depth maps, we effectively distill the teacher's geometric priors into the event-based student. (ii) At the \textit{feature level}, to fully exploit the teacher’s internal representations, we propose \textbf{Spatio-Temporal Feature Distillation (STFD)}, aligning the student’s intermediate features with those of VGGT to enforce the learning of spatial structure and temporal correspondence. 
(iii) At the \textit{temporal level}, we introduce a \textbf{Temporal Consistency Distillation (TCD)} loss to supervise the inter-frame changes of consecutive depth maps rather than their absolute values. By penalizing discrepancies in inter-frame depth changes, TCD compels the student to inherit the teacher's geometrically coherent temporal flow, yielding accurate and stable depth sequences.

We validate EventVGGT on EventScape~\cite{Gehrig2021CombiningEA} and MVSEC~\cite{Zhu2018TheMS} datasets, achieving new state-of-the-art performance against both event- and image-based methods. Furthermore, we demonstrate its robust zero-shot generalization: when trained solely on EventScape and evaluated on the unseen DENSE~\cite{HidalgoCarrio2020LearningMD} and MVSEC datasets, EventVGGT maintains a substantial performance margin. The versatility of our approach is further highlighted by its effective extension to distilling VGGT for other geometric tasks, including camera pose and point map estimation from events.

In summary, our main contributions are three-fold:

\begin{itemize}
    \item We introduce EventVGGT, the first framework to distill spatio-temporal priors from a multi-view foundation model (VGGT) into an event-based student, enabling temporally consistent, annotation-free depth estimation.

    \item We propose a comprehensive tri-level distillation strategy: Cross-Modal Feature Mixture (CMFM), Spatio-Temporal Feature Distillation (STFD), and Temporal Consistency Distillation (TCD).

    \item EventVGGT achieves state-of-the-art results on EventScape and MVSEC, demonstrates strong zero-shot generalization on DENSE, and seamlessly extends to camera pose and point cloud estimation.

\end{itemize}

\section{Related Work}
\label{sec:related_work}
\noindent \textbf{Event-based monocular depth estimation.}
Traditional RGB-based monocular depth estimation~\cite{cao2025panda, cao2025st, yang2024depth} is constrained by fixed frame rates and degrades severely under challenging illumination. To overcome these limitations, event-based depth estimation has emerged as a robust alternative, enabling high-frequency predictions in dynamic lighting and rapid motion scenarios~\cite{zheng2023deep, shi2023even}. Early works, such as E2Depth~\cite{HidalgoCarrio2020LearningMD}, leverage recurrent convolutional networks to generate dense depth maps from event streams. Subsequent methods~\cite{gehrig2021combining, sabater2023event, devulapally2024multi, pan2024srfnet,liu2025CFRNet} integrate paired RGB and event data to further enhance accuracy. Despite these advancements, most supervised approaches remain bottlenecked by their reliance on expensive, extensively annotated datasets. To bypass this, recent annotation-free works like DepthAnyEvent~\cite{bartolomei2025depth} and EventDAM~\cite{zhu2025depth} utilize cross-modal distillation to transfer knowledge from image-based Vision Foundation Models (VFMs). However, by treating event streams as isolated frames, these methods fail to adequately exploit the inherent temporal dynamics of events and the temporal priors of VFMs. \textit{To address this, our work explicitly models the event stream as a continuous video sequence, ensuring that the estimated depth maps maintain strict geometric consistency throughout the sequence}.

\noindent \textbf{Cross-modal knowledge distillation.}  
Knowledge distillation (KD) classically transfers learned representations from a high-capacity teacher to a more compact student network~\cite{wang2021knowledge, hinton2015distilling, zhu2023good, zhu2024clip}. Building upon this, cross-modal KD~\cite{lee2023decomposed, jang2024stxd} extends the paradigm to bridge distinct sensory domains. For instance, early approaches~\cite{gupta2016cross} facilitate representation learning in an unlabeled target modality by exploiting mid-level features from a paired, labeled source. More recently, methods~\cite{zhu2024source, ahmed2022cross} have tackled the challenging setting of transferring knowledge without direct access to task-relevant source data. \textit{In this study, rather than relying on standard feature matching, we propose a comprehensive tri-level distillation strategy (CMFM, STFD, and TCD) to effectively bridge the substantial modality gap between paired RGB sequences and asynchronous event streams}.

\noindent \textbf{Vision foundation models (VFMs).} VFMs have demonstrated remarkable generalization capabilities, driven by large-scale datasets~\cite{kirillov2023segment, radford2021learning, yang2024depth} and advances in self-supervised learning~\cite{caron2021emerging, Oquab2023DINOv2LR, he2022masked}. For example, models like CLIP~\cite{radford2021learning} and SAM~\cite{kirillov2023segment, ravi2024sam} exhibit robust zero-shot transfer across diverse downstream tasks, while the DINO family~\cite{Oquab2023DINOv2LR, simeoni2025dinov3} extracts highly versatile visual representations from massive curated corpora. Building on these breakthroughs, recent efforts have adapted VFMs to alternative modalities, harnessing their representational power for tasks like semantic segmentation~\cite{chen2024segment, kong2024openess} and depth estimation~\cite{zhu2025depth, bartolomei2025depth}. Specifically, EventDAM~\cite{zhu2025depth} employs a dense-to-sparse distillation strategy to bypass the need for dense depth annotations, while DepthAnyEvent~\cite{bartolomei2025depth} utilizes cross-modal distillation to transfer the rich visual priors of image-based VFMs into the event domain. \textit{Advancing this paradigm, we introduce a framework that explicitly distills the rich multi-view 3D priors from a sequence-aware foundation model (VGGT), effectively elevating event-based distillation from the spatial to the spatio-temporal domain.}

\section{Method}
\label{sec:method}
\subsection{Overview}
\begin{figure}[t]
    \centering
    \includegraphics[width=0.95\textwidth]{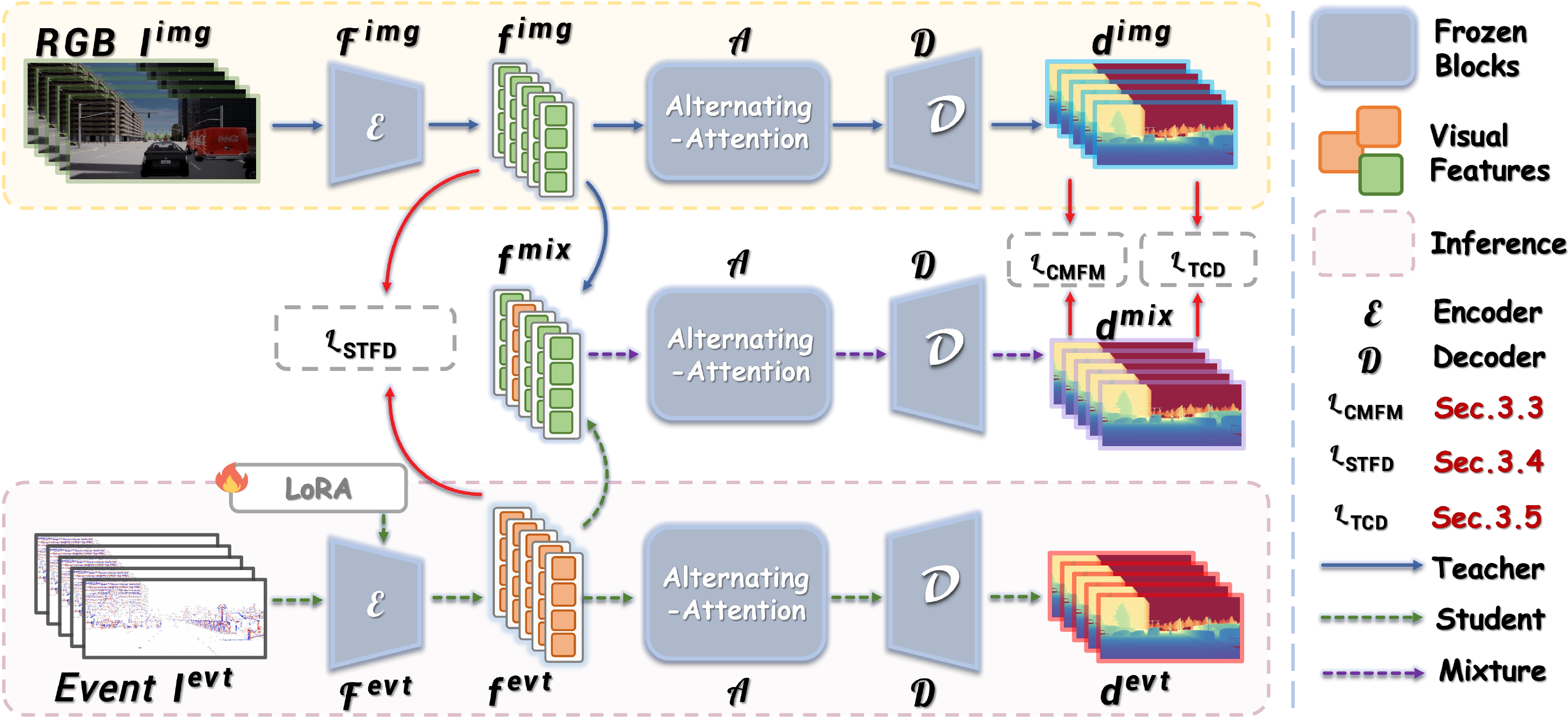}
        \captionof{figure}{\textbf{Overview of the EventVGGT framework.} Our approach distills robust multi-view geometric priors from VGGT to achieve consistent event-based depth estimation. Given synchronized event sequences $I^{evt}$ and RGB images $I^{img}$, we extract modality-specific features using an event encoder $\mathcal{F}^{evt}$ and an image encoder $\mathcal{F}^{img}$, respectively. We introduce three core modules: (i) \textbf{Cross-Modal Feature Mixture (CMFM)}, which constructs mixed feature representations to predict auxiliary depths $d^{mix}$ (\textit{cf.}~Sec.~\ref{sec:cross_modal_feature_mixture}); (ii) \textbf{Spatio-Temporal Feature Distillation (STFD)}, which aligns intra-frame spatial and inter-frame temporal representations. (\textit{cf.}~Sec.~\ref{sec:Spatio_Temporal_Feature_Distillation}); and (iii) \textbf{Temporal Consistency Distillation (TCD)}, which explicitly enforces geometric stability across consecutive depth predictions (\textit{cf.}~Sec.~\ref{sec:temporal_consistency}).}
        \label{fig:framework}
\end{figure}

Our study represents an initial effort to leverage VGGT~\cite{wang2025vggt} to achieve consistent event-based depth estimation, operating entirely without requiring access to ground-truth depth data. Fig.~\ref{fig:framework} illustrates the overall architecture of EventVGGT. We first detail the foundational components—including the inputs, event frame-like representation, and feature encoding—in Sec.~\ref{sec:event_based_mde}. To facilitate robust knowledge transfer from the RGB-based VGGT teacher to the event-based EventVGGT student, we propose a tri-level distillation strategy comprising three core modules: Cross-Modal Feature Mixture (CMFM, Sec.~\ref{sec:cross_modal_feature_mixture}), Spatio-Temporal Feature Distillation (STFD, Sec.~\ref{sec:Spatio_Temporal_Feature_Distillation}), and Temporal Consistency Distillation (TCD, Sec.~\ref{sec:temporal_consistency}).

\subsection{Event-based Monocular Depth Estimation}
\label{sec:event_based_mde}
\noindent\textbf{Inputs.} Unlike conventional frame-based sensors that record absolute intensity values at fixed intervals, event cameras capture visual information as a continuous stream of asynchronous intensity changes. Formally, each event is defined as a tuple $e = (x, y, t, p)$, where $(x, y)$ denotes the pixel coordinates, $t$ is the precise timestamp of the change, and $p \in \{+1, -1\}$ indicates the polarity (representing an increase or decrease in logarithmic intensity). Concurrently, an RGB camera captures standard intensity images, denoted as $I^{img} \in \mathbb{R}^{3 \times H \times W}$, where $H$ and $W$ are the spatial dimensions. Ultimately, the input to the EventVGGT framework comprises a synchronized sequence of $N$ image frames, $I^{img}=\{I^{img}_i\}_{i=1}^N$, alongside the corresponding continuous event stream.

\noindent\textbf{Event frame-like representation.} To facilitate efficient computation and ensure compatibility with standard vision architectures, the continuous event stream is partitioned into fixed temporal windows and accumulated into dense frame-like representations. Following prior studies~\cite{Gehrig2021CombiningEA, pan2024srfnet, HidalgoCarrio2020LearningMD, sabater2022event}, we divide the event stream into discrete temporal windows of $\Delta T = 50$ ms, discretizing each window into $B = 5$ temporal bins. Positive and negative polarity events are subsequently aggregated within each bin to construct histogram-based channels, yielding a synchronized event sequence $I^{\mathrm{evt}} = \{I^{\mathrm{evt}}_i\}_{i=1}^N$, where each frame $I^{\mathrm{evt}}_i \in \mathbb{R}^{3 \times H \times W}$. Finally, these event representations are temporally synchronized with the RGB images, producing perfectly aligned input sequences for the EventVGGT network.

\noindent \textbf{Feature encoding.} Each frame in the RGB sequence $I^{img}$ and the event sequence $I^{evt}$ is divided into $L$ non-overlapping square patches of size $P \times P$, where $L = \frac{H \times W}{P^2}$. The event-based student encoder $\mathcal{F}^{evt}: \mathbb{R}^{3 \times H \times W} \rightarrow \mathbb{R}^{D \times \hat{H} \times \hat{W}}$ processes the event frame sequence to output $D$-dimensional feature maps $f^{evt}=\{f^{evt}_i\}_{i=1}^N$, with reduced spatial dimensions $\hat{H} = \frac{H}{P}$ and $\hat{W} = \frac{W}{P}$. Similarly, the RGB-based teacher encoder $\mathcal{F}^{img}$ produces image feature maps $f^{img}=\{f^{img}_i\}_{i=1}^N$ of identical dimensions. Both encoders share a core network architecture comprising an Alternating-Attention Transformer $\mathcal{A}$ from VGGT~\cite{wang2025vggt}, followed by the depth prediction head. Finally, a depth decoder $\mathcal{D}$ predicts the depth maps $d^{evt}=\{d^{evt}_i\}_{i=1}^N$ and $d^{img}=\{d^{img}_i\}_{i=1}^N$ from the corresponding event and image features, respectively.

\subsection{Cross-Modal Feature Mixture}
\label{sec:cross_modal_feature_mixture}

Due to the substantial modality gap between dense, absolute-intensity RGB images and sparse, asynchronous event streams, directly forcing an event-based student to mimic an RGB teacher's output space often leads to severe gradient conflicts, unstable convergence, and suboptimal adaptation. To overcome this bottleneck, we introduce the Cross-Modal Feature Mixture (\textbf{CMFM}) module at \textit{output level}. Rather than enforcing an overly rigid, direct supervision that hinders the training process, CMFM constructs a ``stepping stone'' to gently bridge the modalities at the output level. By stochastically mixing RGB and event features into a unified sequence and decoding it into an auxiliary prediction, we compel the shared decoder to treat event representations as functionally equivalent to RGB features. This strategy naturally pulls the event stream into the teacher's highly structured feature space, facilitating a much smoother knowledge transfer.

Formally, given the extracted frame-level feature representations ($f^{img}$ and $f^{evt}$) from the synchronized RGB sequence $I^{img}$ and event sequence $I^{evt}$, we synthesize a mixed feature sequence $f^{mix}$. This is achieved by stochastically substituting a subset of the RGB features with their temporally aligned event counterparts. Empirically, we uniformly randomly replace 25\% of the RGB features with event features (e.g., resulting in a sequence like $f^{mix} = \{f^{img}_1, f^{evt}_2, \\f^{img}_3, \dots, f^{img}_N\}$). 
This specific substitution ratio strikes an optimal balance, providing enough RGB context to maintain geometric stability while introducing sufficient event features to enforce cross-modal alignment (comprehensive justification is provided in Tab.~\ref{tab:ablation about mix}).

Once constructed, the mixed feature sequence $f^{mix}$ is fed through the shared Alternating-Attention Transformer $\mathcal{A}$ and depth decoder $\mathcal{D}$ to generate an auxiliary sequence of depth maps, $d^{mix}=\{d^{mix}_i\}_{i=1}^N$. By explicitly supervising this auxiliary output with the high-fidelity depth $d^{img}$ generated by the VGGT teacher, the distillation process is substantially stabilized. 

Following the optimization strategy of VGGT~\cite{wang2025vggt}, our distillation loss incorporates an aleatoric-uncertainty formulation~\cite{novotny2017learning} to intelligently weight the discrepancy between $d^{mix}$ and $d^{img}$ based on the predicted confidence map $\Sigma^{d^{mix}}$. Furthermore, to explicitly preserve local structural consistency---a critical factor for dense monocular depth estimation---we augment this objective with a gradient-based penalty. The objective of CMFM module is formulated as:

\setlength{\abovedisplayskip}{3pt} 
\setlength{\belowdisplayskip}{3pt} 
\begin{equation}
    \mathcal{L}_{\text{CMFM}} = \sum_{i=1}^{N} \Big(
    \|\Sigma_i^{d^{img}} \odot (d^{mix}_i - d^{img}_i)\| +
    \|\Sigma_i^{d^{img}} \odot (\nabla d^{mix}_i - \nabla d^{img}_i)\|
    -\alpha \log \Sigma_i^{d^{img}}
    \Big),
\end{equation}
where $\odot$ denotes the channel-broadcast element-wise product and $\alpha$ is a balancing hyperparameter. Conceptually, $\mathcal{L}_{\text{CMFM}}$ serves as a vital cross-modal bridge. By aligning the mixed-feature auxiliary predictions with the teacher's outputs, it not only stabilizes training but ensures the student model successfully inherits the teacher's robust geometric priors for event-based depth estimation.

\subsection{Spatio-Temporal Feature Distillation}
\label{sec:Spatio_Temporal_Feature_Distillation}

While CMFM successfully bridges the modality gap at the output level, the student network must also inherit the rich geometric priors embedded deep within the VGGT teacher's intermediate representations. However, this presents a fundamental challenge: \textit{event streams intrinsically encode ultra-high-frequency temporal dynamics (motion), whereas the VGGT teacher is optimized to capture spatial scene structures across discrete, static frames.} Prior works, such as EventDAM~\cite{zhu2025depth} and DepthAnyEvent~\cite{bartolomei2025depth}, ignore this discrepancy by processing event streams as independent static frames. This standard frame-by-frame alignment forces dynamic event features into a rigid static representation, destroying their inherent temporal structure and leading to temporally inconsistent and less accurate depth predictions.

To address this discrepancy, we introduce Spatio-Temporal Feature Distillation (\textbf{STFD}) at \textit{feature level}. Rather than merely enforcing a rigid spatial mimicry that suppresses event dynamics, STFD explicitly models and transfers the temporal changes of the features. Formally, given the RGB features $f^{img}$ and corresponding event features $f^{evt}$, our objective is composed of two complementary terms:

\setlength{\abovedisplayskip}{6pt} 
\setlength{\belowdisplayskip}{3pt} 
\begin{equation}
\mathcal{L}_{\mathrm{STFD}} = \sum_{i=1}^{N} \big( 1 - \cos(f_i^{evt}, f_i^{img}) \big) + \sum_{i=1}^{N-1} \big( 1 - \cos(f_{i+1}^{evt}-f_{i}^{evt}, f_{i+1}^{img}-f_{i}^{img}) \big),
\end{equation}
where $\cos(\cdot)$ denotes the channel-wise cosine similarity. The first term performs intra-frame spatial distillation, anchoring the event features to the robust structural geometry embedded within the VGGT backbone. Crucially, the second term executes inter-frame temporal distillation by comparing frame-to-frame feature variations. By explicitly matching the magnitudes and directions of these temporal transitions in the latent space, we ensure that the student network learns motion-sensitive dynamics that are consistent with the RGB teacher's temporal reasoning, fully exploiting the continuous nature of the event stream.

\subsection{Temporal Consistency Distillation}
\label{sec:temporal_consistency}

While CMFM and STFD successfully align the modalities at the spatial output and intermediate feature levels, a significant challenge remains at the final prediction stage. Because event streams capture sparse, asynchronous brightness changes rather than dense photometric data, event-based dense predictions are inherently susceptible to high-frequency temporal instability, often manifesting as severe depth flickering. Standard distillation objectives fail to address this, as they naively penalize per-frame absolute depth errors while ignoring inherent temporal continuity and geometric coherence across frames.

To close this gap, we introduce Temporal Consistency Distillation (\textbf{TCD}) at \textit{temporal level}. The teacher model, VGGT, inherently enforces multi-view geometric consistency across all frames simultaneously via its global feed-forward architecture, producing depth sequences with dynamically smooth, physically realistic transitions. \textit{Our key innovation is to explicitly align the student's inter-frame depth variations with the temporally coherent transitions of the teacher model}. We formulate the TCD objective to penalize discrepancies in the inter-frame rate of change between the student and teacher outputs:

\setlength{\abovedisplayskip}{3pt} 
\setlength{\belowdisplayskip}{3pt} 
\begin{equation}
\mathcal{L}_{\mathrm{TCD}} = \frac{1}{N-1} \sum_{i=1}^{N-1}
\big\|\, | d^{mix}_{i+1} - d^{mix}_i | - | d^{img}_{i+1} - d^{img}_i | \,\big\|_{1}.
\end{equation}

The term $| d^{img}_{i+1} - d^{img}_i |$ calculates the per-pixel magnitude of depth variation across consecutive frames, functioning as a continuous temporal gradient map. This gradient captures the scene's underlying geometric dynamics, including camera ego-motion and dynamic objects. By minimizing the $L_1$ distance between these temporal difference fields, $\mathcal{L}_{\mathrm{TCD}}$ acts as a strict temporal-geometric constraint. It effectively suppresses physically implausible inter-frame discontinuities, anchoring the student's output dynamics to the multi-view-optimized evolution of the VGGT teacher. Ultimately, this guarantees that the student produces spatially precise and dynamically coherent depth sequences.

\noindent \textbf{Total objective.} Combining our proposed distillation components, the final training objective of the EventVGGT framework is formulated as a weighted combination of cross-modal, spatio-temporal, and temporal consistency losses:
\begin{equation}
\mathcal{L} = \mathcal{L}_{\text{CMFM}} + \lambda_{\text{STFD}} \mathcal{L}_{\text{STFD}} + \lambda_{\text{TCD}} \mathcal{L}_{\text{TCD}},
\end{equation}
where $\lambda_{\text{STFD}}$ and $\lambda_{\text{TCD}}$ serve as balancing hyperparameters for the feature-level alignment and temporal consistency distillation, respectively. Empirically, we set $\lambda_{\text{STFD}}$ = 0.1 and $\lambda_{\text{TCD}}$ = 0.2 to effectively regularize the training process.
\section{Experiment}
\label{experiment}
\subsection{Settings}
\textbf{Datasets}. We adopt EventScape~\cite{gehrig2021combining} as our primary synthetic training source due to its large-scale scenes and dense depth annotations. For standard in-domain evaluation, EventVGGT is trained and tested on the respective splits of EventScape and MVSEC~\cite{Zhu2018TheMS}. To assess zero-shot generalization, we train exclusively on EventScape and evaluate directly on the real-world MVSEC and the unseen synthetic DENSE~\cite{HidalgoCarrio2020LearningMD} datasets.


\noindent\textbf{Evaluation metrics}. Following prior works~\cite{pan2024srfnet, devulapally2024multi, hamaguchi2023hierarchical}, we measure average absolute depth error at cut-off distances of 10m, 20m, and 30m. Furthermore, we evaluate our approach using three percentage metrics $\delta_i$, where $i \in \{1.25, 1.25^2, 1.25^3\}$.


\noindent\textbf{Implementation details}. Consistent with prior works~\cite{pan2024srfnet,zhu2025depth}, we center-crop both the RGB images and their corresponding event representations from the EventScape dataset to a spatial resolution of $252 \times 504$ pixels. Aligning with the VGGT training protocol~\cite{wang2025vggt}, sky regions  are masked out to exclude invalid depth values. We optimize the network using AdamW with a learning rate of $1 \times 10^{-4}$. For parameter-efficient fine-tuning~\cite{zhu2025depth}, we apply LoRA ($r = 16$ and $\alpha = 32$), adding approximately 1.6 million trainable parameters. The model is trained on four NVIDIA A800 GPUs with dynamic batch sizes, converging in roughly 11 hours. To effectively distill the rich temporal dynamics from VGGT, we input sequences of 24 synchronized RGB and event frames per batch.

\subsection{Experimental Results}

Following EventDAM~\cite{zhu2025depth}, we evaluate the supervised fine-tuning performance of EventVGGT on EventScape and MVSEC datasets. For the EventScape dataset, the model is trained on training Town 1 subset and evaluated on test set. For MVSEC dataset, EventVGGT is trained on the ``Day2'' subset and subsequently evaluated across the ``Day1'', ``Night1'', ``Night2'', and ``Night3'' subsets. 

\begin{figure}[t]
    \centering
    \includegraphics[width=0.99\linewidth]{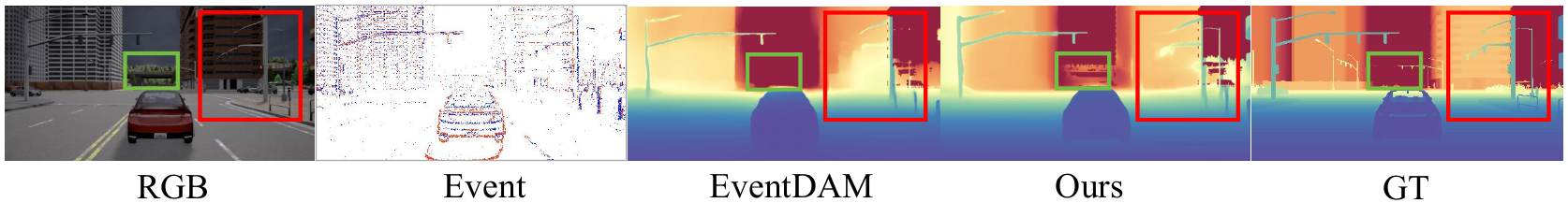}
    \caption{Qualitative results on EventScape dataset.}
    \label{fig:eventscape}
\end{figure}
\noindent \textbf{Comparison on EventScape dataset.} In Tab.~\ref{tab:eventscape1}, we compare the proposed EventVGGT against existing state-of-the-art methods on the EventScape dataset. Overall, EventVGGT consistently outperforms all baselines across evaluated metrics, with the most substantial performance gains observed at greater distances. For instance, while EventDAM yields a depth error of 2.30 at 30m, our method reduces this error to 1.06, representing a 53.9\% relative improvement. 
\begin{wraptable}{r}{0.5\textwidth}
\centering
\caption{Absolute mean depth error results on EventScape (in meters). (E) implies that event data is adopted as the input, and (E+I) means both event and image are adopted. Best results are highlighted in \textcolor{red}{\textbf{Red}} while runner-up in \textcolor{blue}{\textbf{Blue}}.}
\resizebox{0.8\linewidth}{!}{%
\begin{tabular}{l|l|lll}
    \toprule
Method &Input&10m $\downarrow$& 20m $\downarrow$& 30m $\downarrow$\\
\midrule
E2Depth~\cite{HidalgoCarrio2020LearningMD}&E&1.79 &5.35& 8.31\\
\hline
RAMNet~\cite{gehrig2021combining}&\multirow{4}{*}{E+I} & 0.81 &2.26 &3.58\\
HMNet~\cite{hamaguchi2023hierarchical}&&\textcolor{blue}{\textbf{0.55}}& 1.80 &3.27\\
ER-F2D~\cite{devulapally2024multi}&& 0.67 &1.69 &2.81\\
SRFNet~\cite{pan2024srfnet}&& 1.27 &1.68 &2.76\\
\midrule
EventDAM~\cite{zhu2025depth}&\multirow{2}{*}{E} &0.56&\textcolor{blue}{\textbf{1.52}}&\textcolor{blue}{\textbf{2.30}}\\

\textbf{EventVGGT}&&\cellcolor[HTML]{efefef}\textcolor{red}{\textbf{0.54}}&\cellcolor[HTML]{efefef}\textcolor{red}{\textbf{0.79}}&\cellcolor[HTML]{efefef}\textcolor{red}{\textbf{1.06}}\\
    \bottomrule
\end{tabular}}%
\label{tab:eventscape1}
\end{wraptable}
\begin{wrapfigure}{r}{0.5\textwidth}
    \centering
    \includegraphics[width=0.9\linewidth]{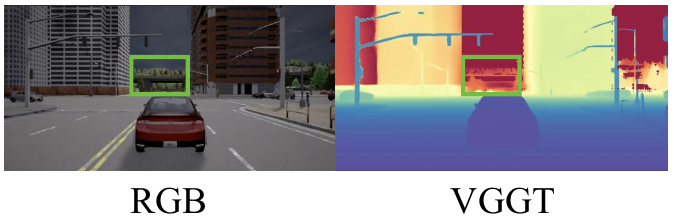}
    \caption{Qualitative result of VGGT on EventScape dataset.}
    \label{fig: vggt}
\end{wrapfigure}
Furthermore, EventVGGT yields lower error rates compared to existing methods that require both event and RGB data (Input: E+I) during inference. For instance, while SRFNet and ER-F2D report depth errors of 2.76 and 2.81 at 30m, respectively, EventVGGT achieves a 1.06 error at the same distance using event data alone. These results demonstrate that EventVGGT learns the robust spatio-temporal representations from the VGGT teacher, effectively bridging the modality gap without requiring auxiliary RGB input at test time.
As shown in Fig.~\ref{fig:eventscape}, visual comparisons on EventScape further validate our quantitative findings. As highlighted by the red bounding boxes, EventVGGT outperforms the EventDAM baseline in recovering fine-grained spatial structures. Our method preserves sharp object boundaries and accurately captures the depth of thin, challenging structures (e.g., streetlights), whereas EventDAM suffers from geometric blurring. Conversely, the green boxes illustrate a specific characteristic of our model: it tends to slightly underestimate the depth of extremely distant background elements, such as far-off buildings. This phenomenon is a direct artifact of the cross-modal distillation process, as VGGT inherently exhibits a similar far-field depth compression bias on this dataset (See Fig.~\ref{fig: vggt}).

\begin{figure}[t]
    \centering
    \includegraphics[width=0.99\linewidth]{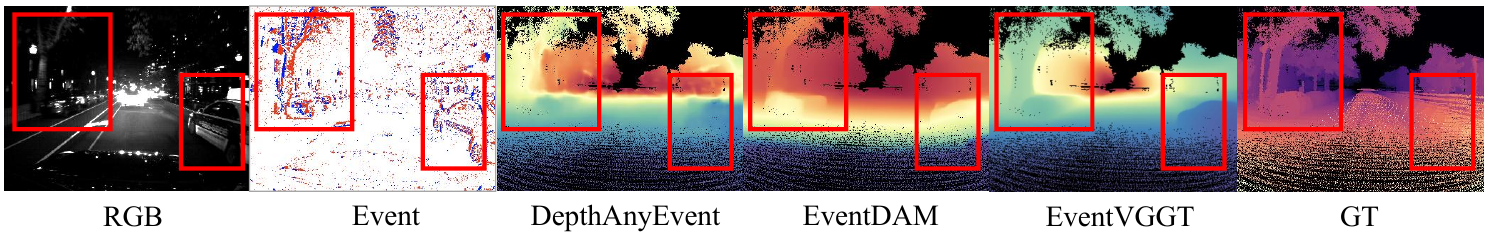}
    \caption{Qualitative results on MVSEC Night1 dataset.}
    \label{fig:mvsec}
\end{figure}

\noindent \textbf{Comparison on MVSEC dataset.} In Tab.~\ref{tab: sub-datasets on MVSEC}, we evaluate EventVGGT on the real-world MVSEC dataset to assess its robustness against extreme illumination changes. Under these challenging conditions, EventVGGT consistently outperforms the event-only EventDAM, reducing the 30m absolute mean depth error from 3.22 to 2.48 on the Night 2 sequence, and from 3.22 to 2.64 on Night 3. While EventDAM's frame-wise distillation suffers from spatial ambiguities when processing erratic nighttime event distributions, EventVGGT overcomes this domain shift by utilizing STFD and TCD to model the continuous video sequence and stabilize predictions using cross-frame dynamics. Furthermore, EventVGGT achieves performance comparable to—and often exceeding—multi-modal (E+I) fusion approaches like SRFNet, which yields a higher 3.52 error at 30m on Night 3 despite relying on paired images. By leveraging the CMFM module 
during daytime training to deeply internalize multi-view geometric priors from the RGB-based VGGT 
teacher, our event-based student model effectively reconstructs missing spatial structures from pure event dynamics, bypassing the need for degraded RGB inputs during inference and mitigating the low-light failure modes of standard cameras. The visual results presented in Fig.~\ref{fig:mvsec} highlight the effectiveness of EventVGGT.

\begin{table}[t]
\centering
\caption{Comparison with methods across varying distances and sub-datasets on MVSEC, evaluated by absolute mean depth error (in meters).}
\resizebox{0.9\linewidth}{!}{%
\begin{tabular}{l|l|lll|lll|lll|lll}
    \toprule
\multirow{2}{*}{Method} & \multirow{2}{*}{Input} & \multicolumn{3}{c|}{Night1} & \multicolumn{3}{c|}{Night2} & \multicolumn{3}{c|}{Night3} & \multicolumn{3}{c}{Day1} \\
\cline{3-14}
& & 10m & 20m & 30m & 10m & 20m & 30m & 10m & 20m & 30m & 10m & 20m & 30m \\
\midrule
E2Depth~\cite{HidalgoCarrio2020LearningMD} & E & 3.38 & 3.82 & 4.46 & 1.67 & 2.63 & 3.58 & 1.42 & 2.33 & 3.18 & 1.67 & 2.64 & 3.13 \\
\hline
RAMNet~\cite{gehrig2021combining} & \multirow{5}{*}{E+I} & 2.50 & 3.19 & 3.82 & \textcolor{blue}{\textbf{1.21}} & 2.31 & 3.28 & \textcolor{red}{\textbf{1.01}} & 2.34 & 3.43 & 1.39 & 2.17 & 2.76 \\
$\text{EvT}^{+}$~\cite{sabater2023event} & & 1.45 &2.10 & 2.88 & 1.48 & \textcolor{blue}{\textbf{2.13}} & 2.90 & 1.38 & \textcolor{blue}{\textbf{2.03}} & 2.77 & 1.24 & 1.91 & \textcolor{blue}{\textbf{2.36}}\\
HMNet~\cite{hamaguchi2023hierarchical} & & 1.50 & 2.48 & 3.19 & 1.36 & 2.25 & 2.96 & 1.27 & 2.17 & 2.86 & 1.22 & 2.21 & 2.68 \\
ER-F2D~\cite{devulapally2024multi} & & 1.58 & 2.24 & \textcolor{blue}{\textbf{2.78}} & 1.54 & 2.23 & 2.95 & \textcolor{blue}{\textbf{1.24}} & \textcolor{red}{\textbf{1.96}} & 2.81 & 1.34 & 2.25 & 2.62 \\
SRFNet~\cite{pan2024srfnet} & & \textcolor{red}{\textbf{1.26}} & \textcolor{red}{\textbf{1.95}} & 3.01 & \textcolor{red}{\textbf{1.19}} & \textcolor{blue}{\textbf{2.13}} & 3.22 & \textcolor{red}{\textbf{1.01}} & 2.12 & 3.52 & \textcolor{red}{\textbf{0.96}} & \textcolor{blue}{\textbf{1.77}} & 2.37 \\
\hline
EReFormer~\cite{EReFormer}&&1.52 &2.28 &2.98 &1.40 &2.12 &2.66 &1.32 &2.04 &\textcolor{blue}{\textbf{2.68}} &1.29 &2.14 &2.59 \\
EventDAM~\cite{zhu2025depth}&\multirow{3}{*}{E}
&\textcolor{blue}{\textbf{1.39}}&2.10&3.25&1.43&2.18&3.22&1.44&2.16&3.22&\textcolor{blue}{\textbf{1.12}}&1.79&2.69\\
Depth AnyEvent~\cite{bartolomei2025depth}&&1.87&2.27&2.81&1.99&2.40&\textcolor{blue}{\textbf{2.86}}&2.05&2.49&2.97&1.50&1.97&2.40\\
\textbf{EventVGGT}&&\cellcolor[HTML]{efefef}1.67&\cellcolor[HTML]{efefef}\textcolor{blue}{\textbf{2.02}}&\cellcolor[HTML]{efefef}\textcolor{red}{\textbf{2.61}}&\cellcolor[HTML]{efefef}1.64&\cellcolor[HTML]{efefef}\textcolor{red}{\textbf{2.03}}&\cellcolor[HTML]{efefef}\textcolor{red}{\textbf{2.48}}&\cellcolor[HTML]{efefef}1.74&\cellcolor[HTML]{efefef}2.15&\cellcolor[HTML]{efefef}\textcolor{red}{\textbf{2.64}}&\cellcolor[HTML]{efefef}\textcolor{red}{\textbf{0.96}}&\cellcolor[HTML]{efefef}\textcolor{red}{\textbf{1.33}}&\cellcolor[HTML]{efefef}\textcolor{red}{\textbf{1.63}}\\
\bottomrule
\end{tabular}}%
\label{tab: sub-datasets on MVSEC}
\end{table}

To further assess the zero-shot capability of our EventVGGT, 
we first train it on the EventScape dataset and subsequently assess its performance on the unseen DENSE and MVSEC datasets.

\noindent \textbf{Comparison on DENSE dataset.} In Tab.~\ref{tab:dense}, we further investigate  the zero-shot generalization capabilities of EventVGGT by training exclusively on the EventScape dataset and directly evaluating on the unseen DENSE dataset. 
\begin{wrapfigure}{r}{0.5\textwidth}
    \centering
    \includegraphics[width=\linewidth]{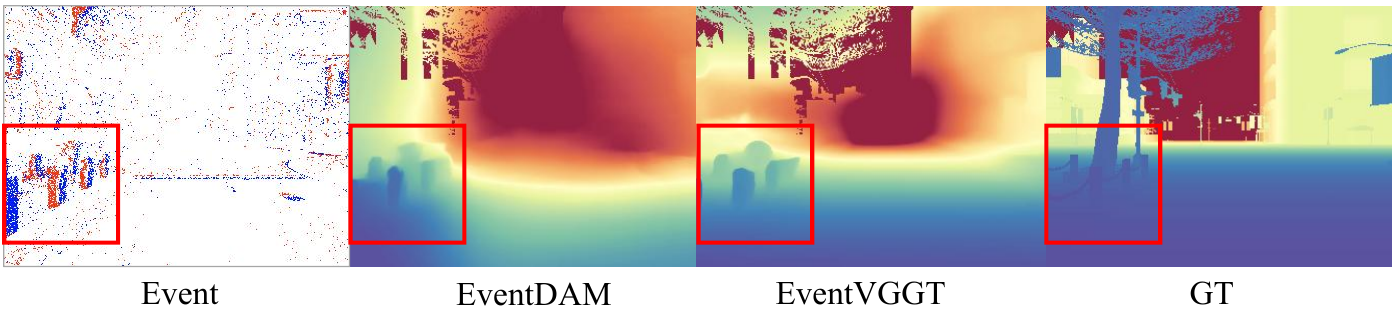}
    \caption{Qualitative results on DENSE.}
    \label{fig: dense}
\end{wrapfigure}
The results highlight EventVGGT's exceptional capacity to transfer learned representations across distinct domains, vastly outperforming multi-modal baselines (Input: E+I) such as RAMNet and SRFNet. And EventVGGT (Input: E) restricts its error to a mere 1.33. Furthermore, EventVGGT substantially outperforms the event-only state-of-the-art EventDAM, which records a higher 5.18 error at 30m. Visual results on DENSE dataset are presented in Fig.~\ref{fig: dense}.

\begin{table*}[t] 
  \centering
  
  \begin{minipage}{0.4\textwidth}
    \centering
    \caption{Absolute mean depth error results on DENSE (in meters).}
    \label{tab:dense}
    \resizebox{\linewidth}{!}{%
      \begin{tabular}{l|l|lll}
        \toprule
        Method &Input&10m $\downarrow$& 20m $\downarrow$& 30m $\downarrow$\\
        \midrule
        RAMNet~\cite{gehrig2021combining}&\multirow{2}{*}{E+I} & 2.62&11.26&19.11\\
        SRFNet~\cite{pan2024srfnet}&& 1.50&3.57&6.12\\
        \midrule
        EventDAM~\cite{zhu2025depth}&\multirow{2}{*}{E} &\textcolor{blue}{\textbf{1.20}}&\textcolor{blue}{\textbf{2.60}}&\textcolor{blue}{\textbf{5.18}}\\
        \textbf{EventVGGT}&&\cellcolor[HTML]{efefef}\textcolor{red}{\textbf{0.54}}&\cellcolor[HTML]{efefef}\textcolor{red}{\textbf{0.89}}&\cellcolor[HTML]{efefef}\textcolor{red}{\textbf{1.33}}\\
        \bottomrule
      \end{tabular}}
  \end{minipage}
  \hfill 
  \begin{minipage}{0.56\textwidth}
    \centering
    \caption{Quantitative results on MVSEC Night1 dataset.}
    \label{tab:comparison with vggt}
    \resizebox{\linewidth}{!}{%
      \begin{tabular}{l|l|llllll}
        \toprule
        Method &Input&10m $\downarrow$& 20m $\downarrow$& 30m $\downarrow$&  $\delta_{1}\uparrow$ & $\delta_{2}\uparrow$ &$\delta_{3}\uparrow$\\
        \midrule
        VGGT~\cite{wang2025vggt}&E&2.42&2.71&3.44&\textcolor{blue}{\textbf{0.42}}&0.69&0.84\\
        VGGT~\cite{wang2025vggt}&I&\textcolor{blue}{\textbf{2.31}}&\textcolor{blue}{\textbf{2.68}}&\textcolor{blue}{\textbf{3.33}}&0.41&\textcolor{blue}{\textbf{0.71}}&\textcolor{blue}{\textbf{0.88}}\\
        \midrule
        \textbf{EventVGGT}&E&\cellcolor[HTML]{efefef}\textcolor{red}{\textbf{1.67}}&\cellcolor[HTML]{efefef}\textcolor{red}{\textbf{2.02}}&\cellcolor[HTML]{efefef}\textcolor{red}{\textbf{2.61}}&\cellcolor[HTML]{efefef}\textcolor{red}{\textbf{0.57}}&\cellcolor[HTML]{efefef}\textcolor{red}{\textbf{0.82}}&\cellcolor[HTML]{efefef}\textcolor{red}{\textbf{0.93}}\\
        \bottomrule
      \end{tabular}}
  \end{minipage}
\end{table*}

\noindent \textbf{Comparison on MVSEC dataset.} We perform a direct zero-shot comparison on the challenging MVSEC Night1 dataset to evaluate the effectiveness of our cross-modal distillation strategy against the original VGGT foundation
 model in Tab.~\ref{tab:comparison with vggt}. When the pre-trained VGGT 
is directly applied to event data (Input: E), it suffers from severe performance degradation due to the massive modality gap and density discrepancies between its native RGB training domain and sparse, asynchronous event streams. By contrast, EventVGGT (Input: E) significantly outperforms the raw VGGT on event data across all distance and $\delta$ accuracy metrics. More impressively, when comparing EventVGGT (Input: E) to the original VGGT operating on its native RGB images (Input: I), our method achieves highly competitive absolute depth errors and demonstrates superior robustness in the $\delta$ metrics. This demonstrates that our distillation pipeline does not merely mimic the teacher network, but effectively empowers the event-based student to leverage the high dynamic range of events in low-light night scenes where standard RGB images severely degrade.

\begin{table*}[t] 
  \centering
  
  \begin{minipage}{0.45\textwidth}
    \centering
    \caption{Effect of each component of EventVGGT on EventScape.}
    \label{tab:ablation_study about component}
\resizebox{\linewidth}{!}{%
\begin{tabular}{lll|lllllll}
\toprule
$\mathcal{L}_{\text{CMFM}}$&$\mathcal{L}_{\text{STFD}}$&$\mathcal{L}_{\text{TCD}}$& 10m $\downarrow$&20m $\downarrow$&30m $\downarrow$ \\
\midrule
\ding{55}&\ding{55}&\ding{55}& 0.69 & 0.94 & 1.26 \\
\ding{51}&\ding{55}&\ding{55}& 0.57 & 0.86 & 1.13\\
\ding{51}&\ding{51}&\ding{55}& 0.53 & 0.81 & 1.10\\
\ding{51}&\ding{55}&\ding{51}& 0.56 & 0.84 & 1.12\\
\ding{51}&\ding{51}&\ding{51}& 0.51 & 0.79 & 1.06\\
\bottomrule
\end{tabular}}%
  \end{minipage}
  \hfill 
  \begin{minipage}{0.5\textwidth}
    \centering
    \caption{Ablation study on the input sequence length.}
    \label{tab:ablation about number}
\resizebox{0.9\linewidth}{!}{%
\begin{tabular}{l|llllll}
\toprule
N &10m $\downarrow$& 20m $\downarrow$& 30m $\downarrow$&  $\delta_{1}\uparrow$ & $\delta_{2}\uparrow$ &$\delta_{3}\uparrow$\\
\midrule
1& 0.79 & 1.10 & 1.50 & 0.82 & 0.96 & 0.99\\
4& 0.75 & 1.05 & 1.41 & 0.83 & 0.97 & 0.99\\
8& 0.70 & 0.98 & 1.31 & 0.85 & 0.97 & 0.99\\
24& 0.71 & 0.95 & 1.26 & 0.85 & 0.97 & 0.99\\
\bottomrule
\end{tabular}}%
  \end{minipage}
\end{table*}

\subsection{Ablation Study}
For all ablation studies, we utilize the training Town 1 dataset of EventScape to train EventVGGT and evaluate the performance with test dataset of EventScape. 

\noindent\textbf{Effect of each component.} As detailed in Tab.~\ref{tab:ablation_study about component}, we validate the individual contributions of our core distillation modules. The naive baseline yields suboptimal depth errors of 0.69, 0.94, and 1.26 at the 10m, 20m, and 30m cut-offs, respectively. Integrating $\mathcal{L}_{\text{CMFM}}$ bridges the cross-modal gap, reducing these errors to 0.57, 0.86, and 1.13. Building on this, $\mathcal{L}_{\text{STFD}}$ explicitly aligns intra-frame geometry and inter-frame dynamics with the teacher model, further decreasing errors to 0.53, 0.81, and 1.10. Finally, the cumulative integration of $\mathcal{L}_{\text{TCD}}$ yields our full EventVGGT framework, achieving minimal errors of 0.51, 0.79, and 1.06. These results validate the necessity of integrating both feature-level alignment and prediction-level temporal consistency for robust cross-modal distillation.

\noindent \textbf{Impact of input sequence length.} In Tab.~\ref{tab:ablation about number}, we ablate the input sequence length to evaluate its impact on extracting spatio-temporal 3D priors from the VGGT teacher. Processing a single frame (N=1) yields suboptimal performance (e.g., a 30m error of 1.50), as the model lacks the multi-view context required for geometric reasoning. Incrementally increasing the sequence length to 4 and 8 frames significantly reduces depth errors and improves $\delta_{1}$ accuracy. Extending the sequence to N=24 frames achieves the best long-range depth estimation, reaching minimum errors of 0.95 and 1.26 at the 20m and 30m cut-offs.
\begin{wraptable}{r}{0.5\textwidth}
\centering
\caption{Ablation study on the CMFM feature mixture rate.}
\resizebox{\linewidth}{!}{%
\begin{tabular}{l|llllll}
\toprule
Rate (\%) &10m $\downarrow$& 20m $\downarrow$& 30m $\downarrow$&  $\delta_{1}\uparrow$ & $\delta_{2}\uparrow$ &$\delta_{3}\uparrow$\\
\midrule
100    & 0.70 & 0.95 & 1.26 & 0.85 & 0.97 & \textbf{0.99}\\
75 & 0.68 & 0.94 & 1.26 & 0.86 & 0.97 & \textbf{0.99}\\
50 & 0.60 & 0.87 & 1.15 & \textbf{0.90} & \textbf{0.98} & \textbf{0.99}\\
25 & \textbf{0.57}& \textbf{0.86} & \textbf{1.13} & \textbf{0.90} & \textbf{0.98} & \textbf{0.99}\\
10  & 0.60 & 0.88 & 1.16 & 0.89 & \textbf{0.98} & \textbf{0.99}\\
\bottomrule
\end{tabular}}%
\label{tab:ablation about mix}
\end{wraptable}
These results support our design choice: processing extended 24-frame sequences fully exploits the temporal multi-view priors of VGGT, providing optimal supervision signals for our cross-modal distillation pipeline.

\noindent \textbf{Influence of CMFM mixture ratio.} Tab.~\ref{tab:ablation about mix} reports the ablation study on the feature replacement rate within the $\mathcal{L}_{CMFM}$
  module. Completely replacing RGB features with event features (100\% rate) yields suboptimal depth estimation, as it removes the dense spatial anchors necessary for stable cross-modal distillation. Lowering the rate to 25\% provides the optimal regularization balance, achieving minimal absolute depth errors (0.57, 0.86, and 1.13 at 10m, 20m, and 30m) and the highest $\delta_1$ 
  accuracy (0.90). This specific ratio preserves sufficient RGB structural context while effectively forcing the event encoder to integrate and compensate for the masked regions. Further reducing the rate to 10\% degrades performance, as the sparse injection of event features weakens the distillation signal. Consequently, we adopt a 25\% replacement rate empirically.

\noindent \textbf{About training dataset.} As reported in Tab.~\ref{tab:ablation about training dataset}, training EventVGGT exclusively on the Town 1 subset yields in-domain performance on EventScape comparable to the model trained on the entire dataset (Towns 1-3). However, it suffers a significant performance drop during zero-shot evaluation on DENSE. This indicates that while expanding synthetic training diversity provides marginal in-domain gains, it is essential for robust cross-domain generalization.

\subsection{Discussion}

\begin{figure}[t]
    \centering
    \includegraphics[width=0.99\linewidth]{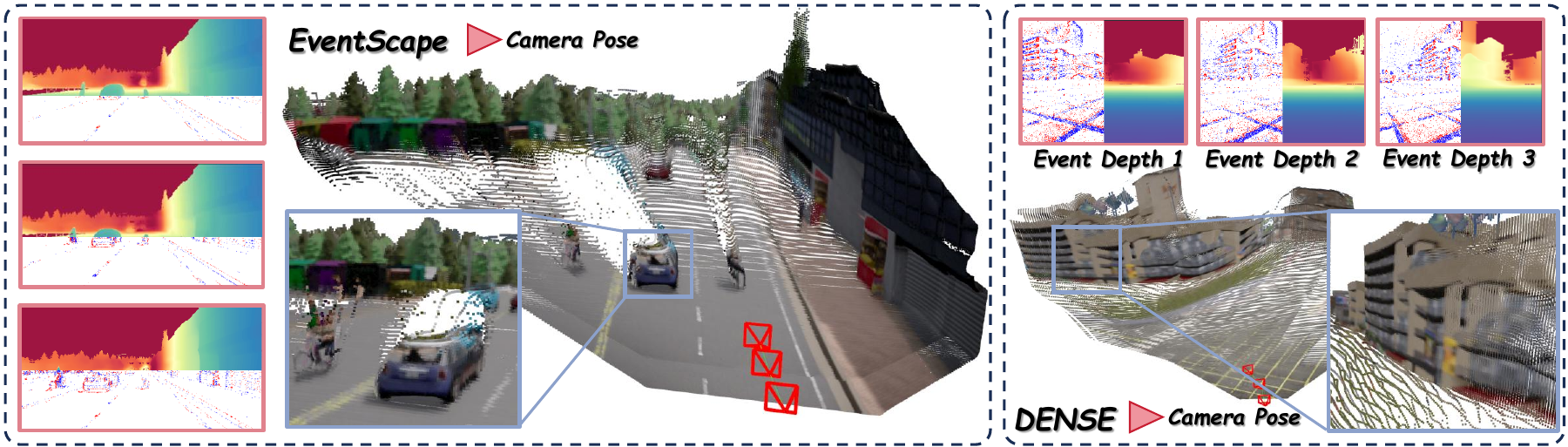}
    \caption{Qualitative results on additional geometric estimation tasks.}
    \label{fig:other}
\end{figure}

\noindent \textbf{Extension to auxiliary 3D tasks.} To demonstrate the versatility of our framework, we extend the distillation pipeline to facilitate camera pose and dense point cloud estimation. Qualitative results on EventScape and DENSE showcase the robust reconstruction of these broader geometric properties (See Fig.~\ref{fig:other}). This highlights the inherent flexibility of the EventVGGT, suggesting that our distilled spatio-temporal priors extend naturally to a broad spectrum of event-based 3D perception tasks beyond monocular depth estimation.

\begin{table}[t] 
\centering
    \caption{Ablation study on the impact of training data diversity.}
\resizebox{0.85\linewidth}{!}{%
\begin{tabular}{l|llllll|llllll}
\toprule
\multirow{2}{*}{Dataset} & \multicolumn{6}{c|}{EventScape} & \multicolumn{6}{c}{DENSE}\\
\cline{2-13}
&10m $\downarrow$& 20m $\downarrow$& 30m $\downarrow$& $\delta_{1}\uparrow$ & $\delta_{2}\uparrow$ &$\delta_{3}\uparrow$& 10m $\downarrow$& 20m $\downarrow$& 30m $\downarrow$&  $\delta_{1}\uparrow$ & $\delta_{2}\uparrow$&$\delta_{3}\uparrow$  \\
\midrule
Town 1& \textbf{0.51} & \textbf{0.79} & 1.09 & \textbf{0.92}& \textbf{0.98} & \textbf{0.99} & 0.64 & 1.03 & 1.50 & 0.87 & \textbf{0.98} & \textbf{0.99}\\
Town 2& 0.61 & 0.86 & 1.16 & 0.89 &\textbf{0.98}& \textbf{0.99}& 0.79 & 1.17 & 1.72 & 0.79 & 0.96 & \textbf{0.99}\\
Town 3& 0.62 & 0.88 & 1.17 & 0.89 & \textbf{0.98} &\textbf{0.99} & 0.82 & 1.24 & 1.75 & 0.74 & 0.96 & \textbf{0.99}\\
Towns 1-3 & 0.53 & \textbf{0.79} & \textbf{1.06}& \textbf{0.92} & \textbf{0.98} &\textbf{0.99} &\textbf{0.54} & \textbf{0.89} &\textbf{1.33} & \textbf{0.87} & \textbf{0.98} & \textbf{0.99}\\
\bottomrule
\end{tabular}}%
\label{tab:ablation about training dataset}
\end{table}

\noindent \textbf{Qualitative analysis of depth consistency.} To validate temporal stability, we back-project inferred multi-frame depths and poses into 3D point clouds. As shown in Fig.~\ref{fig:other}, EventVGGT seamlessly aggregates continuous event streams into unified reconstructions on EventScape and DENSE. By leveraging VGGT's multi-view priors, our framework eliminates the scale inconsistency of frame-by-frame methods, yielding accurate camera extrinsics (red frustums) and precise geometric point cloud alignment. This high-fidelity reconstruction underscores our superiority in consistent event-based depth estimation.

\noindent\textbf{Computational complexity.} By employing LoRA for adaptation, the total parameters of EventVGGT remain highly comparable to that of VGGT~\cite{wang2025vggt}. Correspondingly, EventVGGT demonstrates high inference efficiency, processing a 252$\times$504 event representation in 24 ms on a single NVIDIA A800 GPU.

\section{Conclusion}
\label{limitations}
\noindent We presented EventVGGT, a novel framework for temporally consistent, annotation-free event-based depth estimation. By explicitly modeling asynchronous event streams as continuous video sequences, our approach enables the distillation of robust spatio-temporal and multi-view geometric priors from the image-based VGGT to event domain. To effectively bridge the modality gap, we introduce a comprehensive cross-modal distillation strategy comprising Cross-Modal Feature Mixture (CMFM), Spatio-Temporal Feature Distillation (STFD), and Temporal Consistency Distillation (TCD). Extensive experiments demonstrate that EventVGGT establishes a new state-of-the-art on EventScape and MVSEC, exhibiting robust zero-shot generalization to unseen domains. Furthermore, our framework seamlessly extends to holistic 3D perception tasks, including camera pose and point cloud estimation.

\noindent \textbf{Limitations and Future Work.} EventVGGT inherits a minor far-field depth compression bias from the VGGT teacher model, occasionally leading to the underestimation of extremely distant backgrounds. Future work will integrate dense ground-truth depth into the distillation pipeline to explicitly calibrate this long-range artifact and enhance real-world reliability.

\section{Acknowledgments}

This work was supported in part by the National Natural Science Foundation of China (Grant No. 92370204), in part by the National Key R\&D Program of China(Grant No. 2023YFF0725001), in part by Guangdong Provincial Key Laboratory of Frontier Basic Science for All‑domain Intelligence, in part by the guangdong Basic and Applied Basic Research Foundation (Grant No. 2023B1515120057), in part by the Key-Area Special Project of Guangdong Provincial Ordinary Universities (Grant No. 2024ZDZX1007), in part by Young Scientists Fund of the National Natural Science Foundation of China (Grant No. 42401567), in part by Guangdong Provincial Project (Grant No. 2024QN11G095), in part by The Open Fund of the State Key Laboratory of Spatial Datum (Grant No. SKLSD2026-KF-25), National Natural Science Foundation of China (Grant No. 42571390), in part by Science and Technology Projects of Xizang Autonomous Region, China(Grant No. XZ202501ZY0091), in part by Basic and Applied Basic Research Foundation of Guangdong Province (Grant No. 2025A1515011807), and in part by the Research Grants Council (RGC) of Hong Kong SAR (Grant Nos. GRF14213125, GRF14201824, and GRF14216222).

\clearpage
%
%
\bibliographystyle{splncs04}
\bibliography{main}
\end{document}